\documentclass[letterpaper,10pt,conference]{ieeeconf}

\IEEEoverridecommandlockouts
\let\labelindent\relax

\usepackage{url}

\usepackage{color}
\usepackage{amsmath}
\usepackage{todonotes}
\usepackage{xspace}
\usepackage{enumitem}
\usepackage{algorithm}
\usepackage{algpseudocode}
\usepackage{amsfonts}
\usepackage{cite}
\usepackage{subcaption}
\usepackage[hidelinks]{hyperref}
\usepackage{nameref}
\usepackage{float}
\usepackage{multirow}

\algnewcommand\Or{\textbf{or}}

\newcommand{\revone}[1]{{\color{black}{#1}}}

\newtheorem{proof}{Proof}

\graphicspath{{figs/}}

\begin{document}

\title{\LARGE \bf GATSBI: An Online GTSP-Based Algorithm\\ for Targeted Surface Bridge Inspection}

\author{Harnaik Dhami, Kevin Yu, Troi Williams, Vineeth Vajipey, and Pratap Tokekar
\thanks{Dhami, Williams, Vajipey, and Tokekar are with the Department of Computer Science \& Engineering, University of Maryland, U.S.A. \texttt{\small \{dhami, troiw, vvajipey, tokekar\}@umd.edu}. Yu was with the Department of Electrical \& Computer Engineering, Virginia Tech, U.S.A. Email: \texttt{\small klyu@vt.edu}.}\thanks{This work is supported by the National Science Foundation under grant number 1840044.}}

\maketitle
\IEEEpeerreviewmaketitle

\begin{abstract}
We study the problem of visual surface inspection of a bridge for defects using an Unmanned Aerial Vehicle (UAV). We do not assume that the geometric model of the bridge is known beforehand. Our planner, termed \emph{GATSBI}, plans a path in a receding horizon fashion to inspect all points on the surface of the bridge. The input to GATSBI consists of a 3D occupancy map created online with LiDAR scans. Occupied voxels corresponding to the bridge in this map are semantically segmented and used to create a bridge-only occupancy map. Inspecting a bridge voxel requires the UAV to take images from a desired viewing angle and distance. We then create a Generalized Traveling Salesperson Problem (GTSP) instance to cluster candidate viewpoints for inspecting the bridge voxels and use an off-the-shelf GTSP solver to find the optimal path for the given instance. As the algorithm sees more parts of the environment over time, it replans the path to inspect novel parts of the bridge while avoiding obstacles. We evaluate the performance of our algorithm through high-fidelity simulations conducted in AirSim and real-world experiments. We compare the performance of GATSBI with a classical exploration algorithm. Our evaluation reveals that targeting the inspection to only the segmented bridge voxels and planning carefully using a GTSP solver leads to a more efficient and thorough inspection than the baseline algorithm.
\end{abstract}

\section{Introduction}

In this work, we are interested in designing a high-level planner that inspects a 3D surface, i.e., a bridge for identifying visual defects. Inspection is closely related to coverage and exploration, which are problems that have been well-studied in the literature. However, as we will show, coverage and exploration are not necessarily the best approaches for inspection. Given a 3D model of the environment (including the bridge), we can find a coverage path that covers all points on the bridge using an offline planner~\cite{9048979}. In practice, we often do not have any prior model of the layout of the bridges. Even if a prior 3D model is available, it may be inaccurate due to changes in the environment surrounding the bridge as well as structural changes made to the bridge. In this work, we address the problem of designing targeted inspection plans as the 3D model of the environment is built online. 

Recently, a number of commercial solutions such as the ones from Skydio~\cite{skydio} and Exyn~\cite{Exyn_Technologies_undated-gq} and ongoing work in academia provide robust autonomy including SLAM and low-level planning (how to navigate from point A to point B). Our work on high-level planning (determining what the next waypoint B should be) is complementary to these works. Current forms of planning mostly consist of someone clicking on waypoints for the UAVs to fly to. As a result, we develop tools that autonomously solve the more general problem of inspecting a bridge with no prior information about its geometry.

\begin{figure}
    \centering
    \includegraphics[width = 0.75\columnwidth]{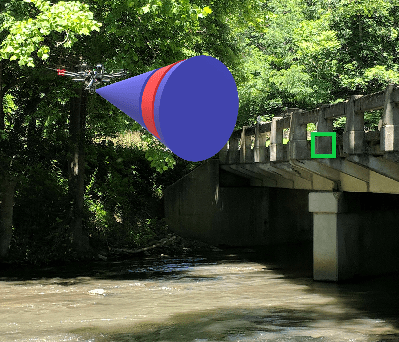}
    \caption{An example of $View$. The green box depicts the face of a bridge voxel, the blue cone depicts the viewing cone, and the red band on the cone depicts the viewing distance.}
    \label{fig:vcvd}
\end{figure}

Frontier-based strategies~\cite{yamauchi1997frontier} are typically used for exploring an initially unknown environment. A frontier is a boundary between explored and unexplored regions. The strategy chooses which of the frontiers to visit (and which path to follow to get to the chosen frontier) to help speed up exploration. The algorithm terminates when there are no more accessible unexplored regions. A bounding box placed around the bridge can restrict exploration when operating in an open environment. Exploring the bridge does not necessarily mean that the UAV will get inspection-quality images. Instead, an inspection planner that can take into account viewing and distance constraints may be more efficient. We present such a planner, termed \emph{GTSP-Based Algorithm for Targeted Surface Bridge Inspection (GATSBI)}, and show that it outperforms the frontier-based exploration strategies in efficiently inspecting the bridge (Section~\ref{sec:eval}).

GATSBI consists of four modules: 3D occupancy grid mapping using LiDAR data, a semantic segmentation algorithm to find LiDAR points that correspond to the surface of the bridge, a Generalized Traveling Salesperson Problem (GTSP) solver for finding inspection paths for the UAV, and a navigation algorithm for executing the planned path. We use off-the-shelf modules for occupancy grid mapping (OctoMap~\cite{hornung2013octomap}), solving GTSP (GLNS~\cite{Smith2016GLNS}), and point-to-point navigation (MoveIt~\cite{coleman2014reducing}). Our key algorithmic contribution is to show \emph{how} to reduce the inspection problem to a GTSP instance and the full pipeline that outperforms baseline strategies. Our technique takes into account overlapping viewpoints that can view the same parts of the bridge and simultaneously selects \emph{where} to take images and \emph{what order} to visit those viewpoints. We also show when and how to replan as we obtain more information about the environment.

\section{Related Work}\label{sec:related}
As described earlier, frontier-based exploration is a widely-used method for 3D exploration of unknown environments~\cite{zhu20153d,da2020novel,niroui2017robot}. Other works proposed variants of frontier exploration focused on choosing the next frontier to visit~\cite{dai2020fast,shen2012autonomous}. Another popular approach is to model the exploration problem as one of information gathering and choose a path (or a frontier) that maximizes the information gain~\cite{corah2019communication,premkumar2020combining}. Additionally, there are Next-Best-View approaches~\cite{pito1999solution} that, as the name suggests, plan the next-best location to take an image from to explore the environment. We refer the reader to a recent, comprehensive survey on multi-robot exploration by Li~\cite{li2020exploration} that covers a variety of exploration strategies.

As shown in our simulations, generic exploration strategies are inefficient when performing targeted inspection (e.g., bridge inspection). There has been work on designing inspection algorithms that plan paths that take into account the viewpoint considerations~\cite{peng2019adaptive,hollinger2013active,roberts2017submodular,song2020online}. When prior information is available, one can plan inspection paths carefully by considering the geometric model of the environment. Typically, algorithms use prior information, such as a low-resolution version of the environment, to create an inspection path and obtain high-resolution measurements of the environment~\cite{peng2019adaptive,roberts2017submodular}. Unlike these works, we consider a scenario where the robot has no prior environmental information and must plan using incrementally revealed data.

Bircher et al.~\cite{bircher2018receding} presented a receding horizon planner for exploration and inspection. Both algorithms use a Rapidly-Exploring Random Tree to generate a set of candidate paths in the known, free space of the environment. Then the algorithm selects a path based on a criterion that values how much information a path gains about the environment. The planner uses a receding horizon algorithm repeatedly invoked with new information. We follow a similar approach; however, their algorithm knows the inspection surface a priori, \revone{while our work only requires the UAV to see a portion of the bridge at the start of inspection}. The environment is not known for their exploration algorithm but it is known for their inspection algorithm. GATSBI has no prior knowledge about the environment and \revone{minimal knowledge of the} target inspection surface. Furthermore, we cluster potential viewpoints using GTSP which leads to further efficiency.

Song et al.~\cite{song2020online} recently proposed an online algorithm that consists of a high-level coverage planner and a low-level inspection planner. The low-level planner takes into account the viewpoint constraints and chooses a local path that gains additional information about the structure under inspection. Our work differentiates by guaranteeing the quality of inspection, not requiring a bounding box around the target infrastructure, and segmenting the infrastructure of interest from the environment which guarantees inspection of only the target infrastructure.

In summary, we make the following contributions:
\begin{itemize}
    \item Present a receding horizon algorithm, GATSBI, that plans paths to efficiently inspect bridges when no prior information about them is present;
    \item Demonstrate that GATSBI outperforms a baseline frontier-based exploration 15x at bridge surface inspection and 17x at crack detection through numerous simulations in AirSim using 3D models of bridges;
    \item Validate the practical feasibility with experiments on a mock bridge and a simplified version of GATSBI;
    \item Provide ROS packages that integrate MoveIt with AirSim for simulations and DJI's SDK for real-world experiments.\footnote{\url{https://github.com/raaslab/GATSBI}}
\end{itemize}

\section{Problem Formulation}\label{sec:prob}
We describe the overarching problem below.
\paragraph*{Problem 1}\label{prob:1}
Given a UAV with a 3D LiDAR and RGB camera, find a path to inspect every point on the bridge surface to minimize the total flight distance.

We consider the scenario where the geometric model of the bridge is unknown a priori. We assume that the UAV starts the algorithm at a location where at least some part of the bridge is visible. If this is not the case, we can run a frontier exploration strategy until the bridge is visible. We then plan an inspection path for the part of the bridge that is visible. As the UAV sees more of the bridge, we replan to find a better tour in a receding horizon fashion. ~\nameref{prob:2} (defined shortly) focuses on finding a path for the UAV based on the partial information. We solve the larger~\nameref{prob:1} by repeatedly solving~\nameref{prob:2} as we gain new information. 

We use a 3D semantic, occupancy grid built using localized pointcloud data from the 3D LiDAR to represent the model of the bridge built online. GATSBI assigns each voxel in the occupancy grid a semantic label. The label indicates whether the voxel is free space $v_{F} \in V_{F}$; is occupied space, part of the bridge, and previously inspected $v_{BI} \in V_{BI}$; is occupied, part of the bridge, but not yet inspected $v_{BN} \in V_{BN}$; and occupied but not a bridge voxel (i.e., obstacles)  $v_{O} \in V_{O}$. Our goal is to inspect all the voxels that correspond to the bridge surface, i.e., to ensure that $V_{BN} = \emptyset$. 

\begin{figure}
    \centering
    \includegraphics[width = 0.85\columnwidth]{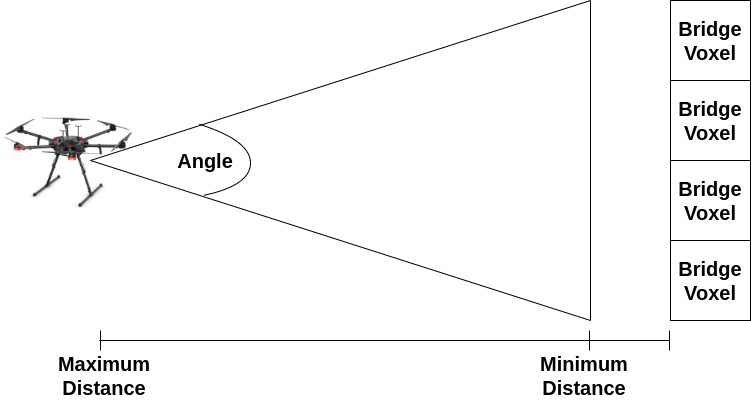}%
    \caption{Example of $View$ where bridge voxels can be inspected.}%
    \label{fig:inspectionExample}%
\end{figure}

A voxel $v_{BN} \in V_{BN}$ is inspected if we inspect at least one of its six faces. A face is inspected if the center of that face falls within a cone given apex angle centered at the UAV camera and within a minimum and maximum range of the UAV camera. The apex angle represents the field of view of the camera that is rigidly attached to the UAV. The viewing distance is a minimum and maximum distance range that the UAV should inspect a bridge voxel to ensure quality images for inspection. Figure~\ref{fig:inspectionExample} shows an example of these viewing constraints. 
%Figure~\ref{fig:vcvd} shows an example of viewing constraints where the green box represents a face of a bridge voxel, the blue cone represents the viewing cone, and the red band on the cone represents the viewing distance. 
For the rest of the paper, we refer to the viewing cone and distance as $View$. The RGB camera is used to take pictures of the bridge once the UAV has reached a target $View$ point.

\paragraph*{Problem 2} \label{prob:2}
Given a 3D occupancy map consisting of four sets of voxels ($V_{F}, V_{BI}, V_{BN}, V_{O}$), find a minimum length path that inspects every voxel in $V_{BN}$.

In the following section, we show how to model this problem as a GTSP instance.

\section{The GATSBI Planner} \label{sec:GATSBI}
\begin{figure}
    \centering
    \includegraphics[width = 0.7\columnwidth]{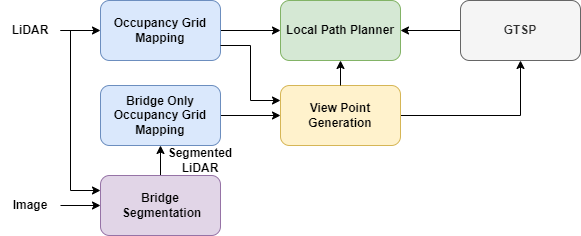}
    \caption{Flow diagram of GATSBI. \revone{The algorithm creates an occupancy map of the environment using incoming LiDAR scans. Then, it segments the points corresponding to the bridge into another point cloud using the RGB camera images. It then makes another occupancy map of only the bridge using the segmented point cloud. GATSBI uses both the environment and bridge occupancy maps to generate viewpoints, points in free space where the UAV can inspect the bridge. It sends these to the GTSP instance to make a tour and then a local path planner to get the flight path.}}
    \label{fig:flowDiagram}
\end{figure}

In this section, we give an overview of the GATSBI algorithm. We show the full pipeline (Fig.~\ref{fig:flowDiagram}) which broadly consists of two modules: perception and planning. We describe each in detail next.

\subsection{Perception} \label{sec:perception}

\begin{figure}
    \centering
    \includegraphics[width = 0.85\columnwidth]{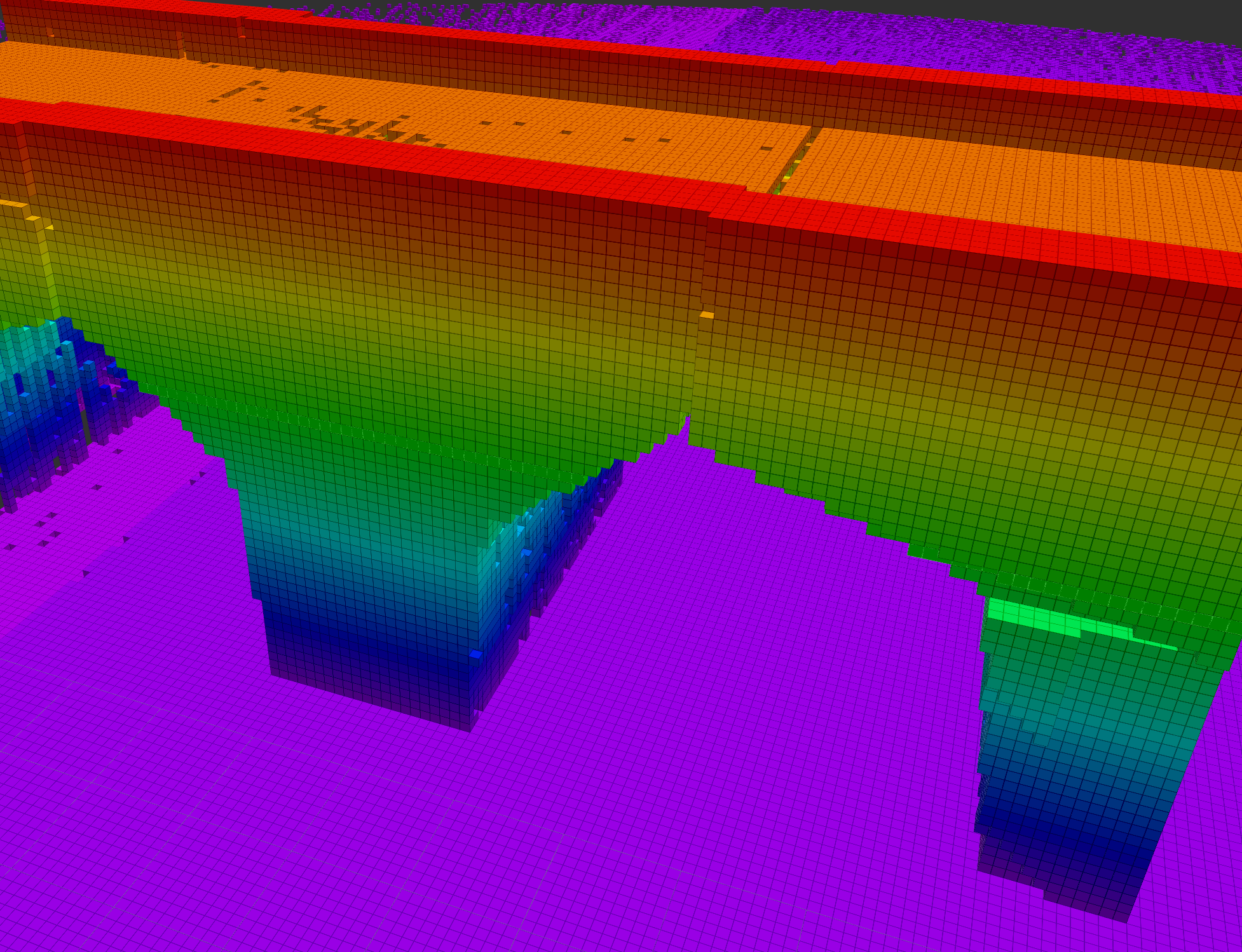}%
    \caption{Full voxel map containing $V_{BI} \cup V_{BN} \cup V_O$.}%
    \label{fig:fullAndSeg}%
\end{figure}

GATSBI starts by segmenting the bridge from the environment. Segmenting out the bridge allows the algorithm to differentiate between the bridge and obstacles in the environment. This allows only the bridge to be inspected as opposed to every object in the environment. The 3D LiDAR points that lie on the segmented bridge are classified as bridge points. These bridge points are then used to create a 3D occupancy grid. In parallel, the complete point cloud (segmented and non-segmented points), is used to generate an environmental 3D occupancy grid. Together, these two occupancy grids output a set of voxels: free $V_F$, bridge $V_{BI}$, bridge $V_{BN}$, and obstacle $V_O$. The algorithm uses the segmented voxels ($V_{BI}, V_{BN}$) to plan inspection paths. The algorithm uses the other voxels ($V_{O}$, $V_{F}$) to plan collision-free paths and take into account viewing constraints. OctoMap~\cite{hornung2013octomap} is used to generate our 3D occupancy grids. An example of the  3D occupancy grid is shown in Fig.~\ref{fig:fullAndSeg}.

\subsection{Planner} \label{sec:planner}

To inspect a bridge, we need to inspect all voxels in $V_{BN}$ (as described in Section~\ref{sec:prob}). GATSBI works in a receding horizon fashion. The $V_{BI}$ set keeps track of inspected voxels. This avoids unnecessarily inspecting the same voxel more than once. Specifically, the UAV must view each voxel in $V_{BN}$ from some point on its path within $View$. We formulate this problem as a Generalized Traveling Salesperson Problem (GTSP) instance. GTSP generalizes the Traveling Salesperson Problem and is NP-Hard~\cite{Smith2016GLNS}. The input to GTSP consists of a weighted graph, $G$, where vertices are clustered into sets.  The edges of the graph are the distances between the vertices. The objective is a minimum weight tour that visits at least one vertex in each set once. Our GTSP configuration is explained next.

For our implementation, vertices are the center-points of voxels $v_F$ that the UAV can fly to and clusters are the set of all $v_F$ a specific $v_{BN}$ can be inspected at. Each vertex in $G$ corresponds to a candidate viewpoint. We check all pairs of $v_F \in V_F$ and $v_{BN} \in V_{BN}$ to see if $v_F$ lies within $View$ of one of the faces of $v_{BN}$. If so, we add a vertex in the graph $G$ corresponding to the pair $v_F$ and $v_{BN}$. 

\begin{figure}
    \centering
    \includegraphics[width = 0.7\columnwidth]{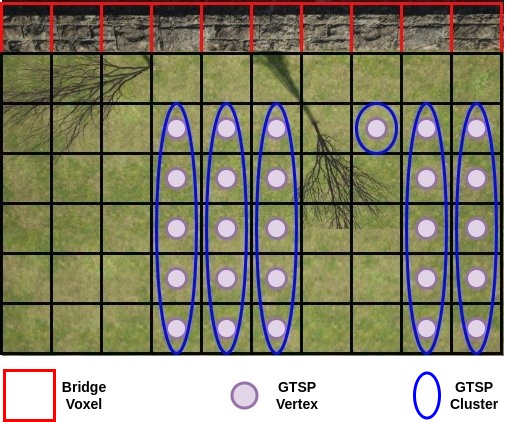}%
    \caption{Example GTSP setup. Each inspectable bridge voxel can have multiple potential inspection viewpoints (vertices). All the vertices for a single bridge voxel are clustered together. The edges between these vertices are initially their Euclidean distance.}%
    \label{fig:gtsp_example}%
\end{figure}

Each free voxel that can inspect the same $v_{BN}$ will add one vertex each to the cluster corresponding to $v_{BN}$. A simplified example of this graph setup is shown in Fig.~\ref{fig:gtsp_example}. The GTSP tour will ensure the UAV visits at least one viewpoint in this cluster.

Next, we create an edge between every pair of vertices in $G$. The cost for each of these edges is initially the Euclidean distance between the two vertices.  With the vertices, edges, and clusters, we create a GTSP instance and use the GTSP solver, GLNS~\cite{Smith2016GLNS}, to find a path for the UAV. We use GTSP as our planner because it will guarantee at least one point corresponding to every $V_{BN}$ will be visited.

Before moving from one point to the next, we check the distance from the current location of the UAV to the next vertex in the GTSP path. Instead of Euclidean distance, we find the distance of the path between these points using a Rapidly-exploring random tree (RRT) connect algorithm with our environment occupancy grid. This ensures the path between these points is collision-free. We use RRT Connect to quickly find a path within a threshold as opposed to using another RRT variant such as RRT* to find the optimal path. If the difference between the RRT distance and the Euclidean distance is greater than $DD$ (discrepancy distance), we update the edge costs in the GTSP instance and replan the GTSP path. We replan as needed to ensure that the first edge in the returned tour is within $DD$ of the Euclidean distance. We call this a lazy evaluation of edge costs. Computing the RRT distance (which is a more accurate approximation of the actual travel distances) between every pair of vertices would be time-consuming. By checking the discrepancy lazily, we find a tour quickly while also not executing any edge where the actual distance is significantly larger than the expected distance. For our experiments, we set $DD$ to be 125\% of the Euclidean distance to account for some of the variances in paths generated using RRT algorithms but still allow replanning when necessary.

\begin{algorithm}
\caption{Overview of single iteration of GATSBI algorithm}\label{alg:gatsbi}
\begin{algorithmic}[1]
\State{Update occupancy grids with latest localized pointcloud as described in Section~\ref{sec:perception}}
\State{Find all inspectable bridge voxels using occupancy grids and remove previously inspected bridge voxels, resulting in $V_{BN}$}
\If{$V_{BN}$ = 0}
\State{Terminate}
\EndIf
\State{Create GTSP instance G as described in Section~\ref{sec:planner}}
\While{Difference in the RRT distance of the first edge in the GTSP solution and its Euclidean distance is greater than a threshold}
\State{Update first edge cost in G with RRT distance and re-solve GTSP}
\EndWhile
\State{Use RRT as point-to-point planner for GTSP tour}
\State{Update $V_{BI}$ with latest inspected bridge voxels}
\end{algorithmic}
\end{algorithm}

We keep track of each newly visited cluster during the flight. Each of these newly visited clusters corresponds to a non-inspected bridge voxel. The camera is also used to take an image at each visited point in the path to obtain inspection images. Once inspected, GATSBI moves them from set $V_{BN}$ to $V_{BI}$. We execute the plan until one of two conditions is met: either a time limit ($RPT$) elapses, or we complete the path, whichever occurs first. We also record the raw sensor data during the flight. Once we complete navigation, we use the stored data to update the bridge inspected and non-inspected voxels and replan. Once $V_{BN}$ is empty, GATSBI considers the bridge inspected and terminates. An overview of a single iteration of the GATSBI algorithm can be viewed in Algorithm~\ref{alg:gatsbi}.

\section{Evaluation}\label{sec:eval}
In this section, we evaluate the algorithm in both simulation and hardware experiments. For the simulations, we compare it against a baseline algorithm as well as discuss parameter tuning.

\subsection{Simulations}\label{sec:sim}
We present simulation results to evaluate the performance of GATSBI. We first present a qualitative example of the inspection paths produced by GATSBI. We also evaluate the computational time for three subroutines within GATSBI. Finally, we compare GATSBI with the baseline frontier-based exploration. 

\begin{figure}
    \centering
    \begin{subfigure}[b]{0.49\columnwidth}
        \includegraphics[width = \textwidth]{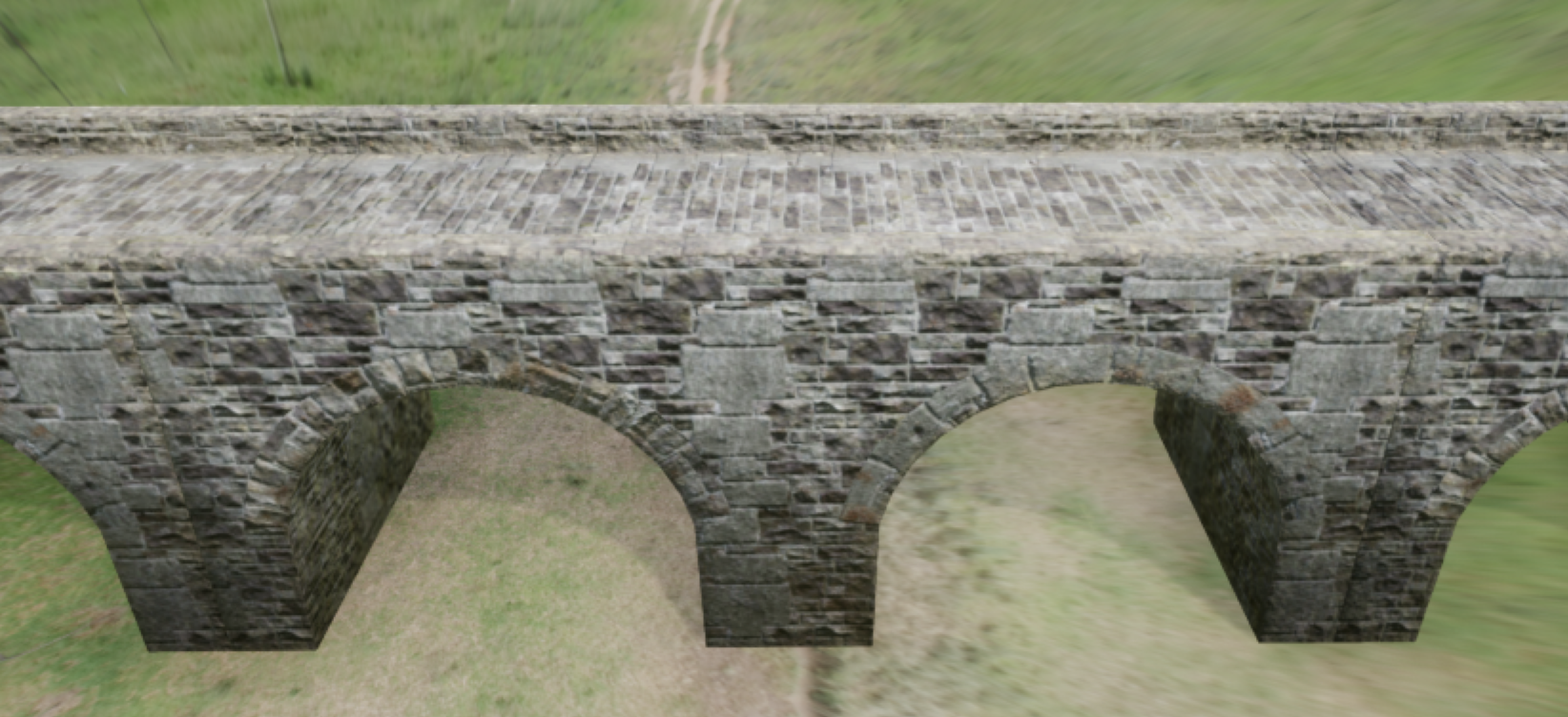}
        \caption{Arch Bridge}
    \end{subfigure}
    \begin{subfigure}[b]{0.49\columnwidth}
        \includegraphics[width = \textwidth]{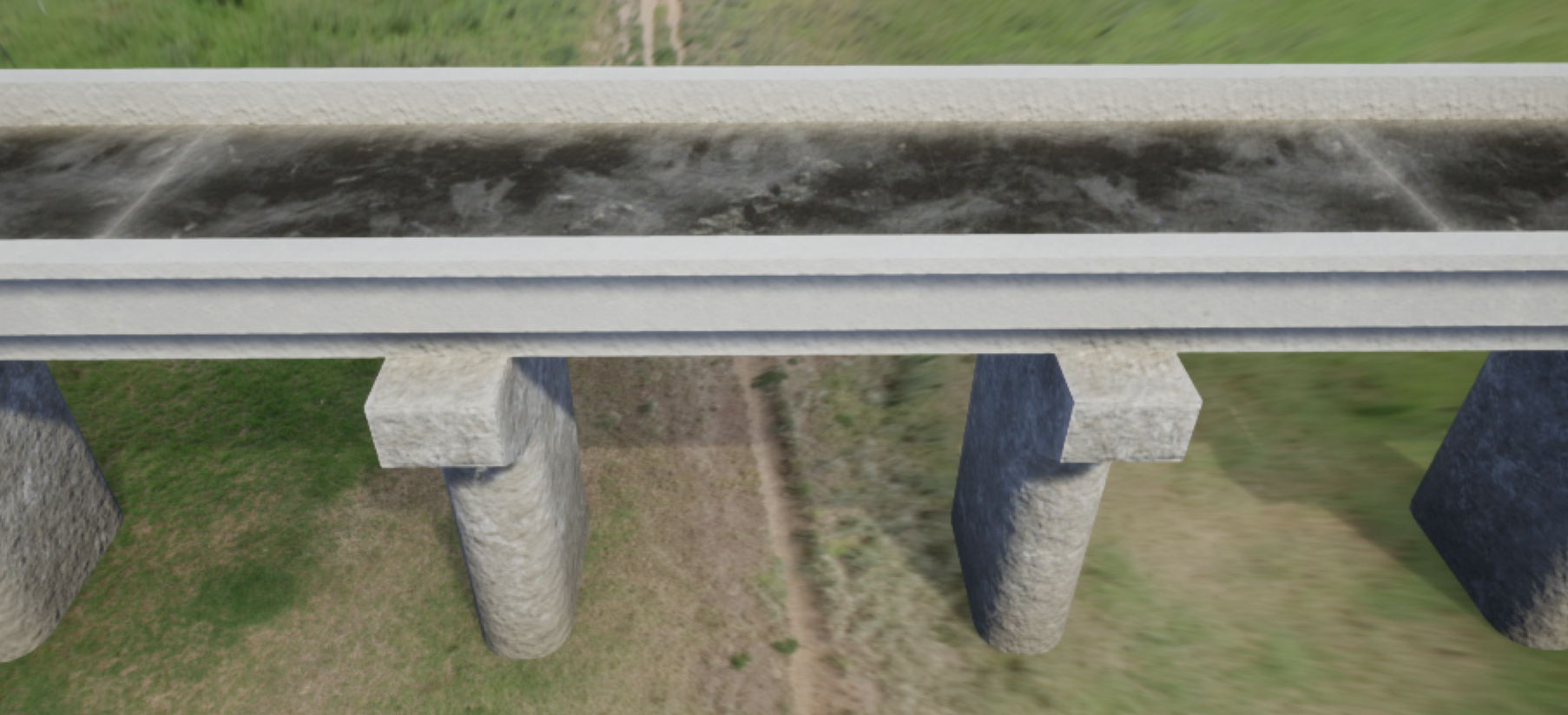}
        \caption{Box Girder Bridge}
    \end{subfigure}
     \begin{subfigure}[b]{0.49\columnwidth}
        \includegraphics[width = \textwidth]{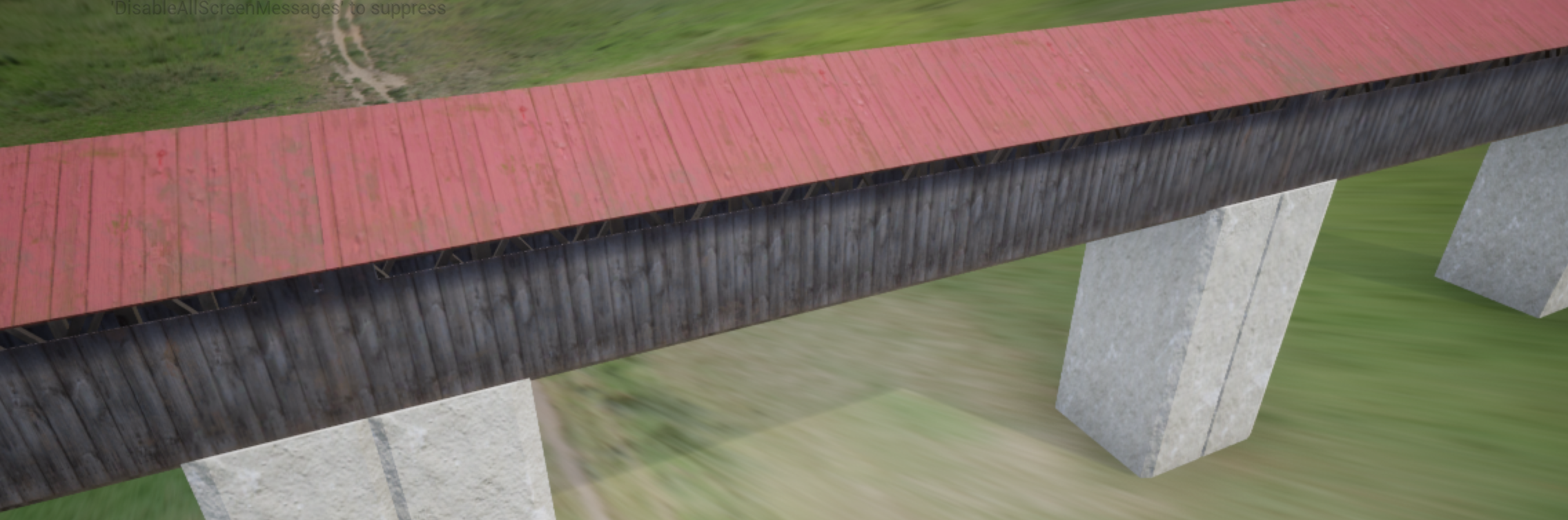}
        \caption{Covered Bridge}
    \end{subfigure}
    \begin{subfigure}[b]{0.49\columnwidth}
        \includegraphics[width = \textwidth]{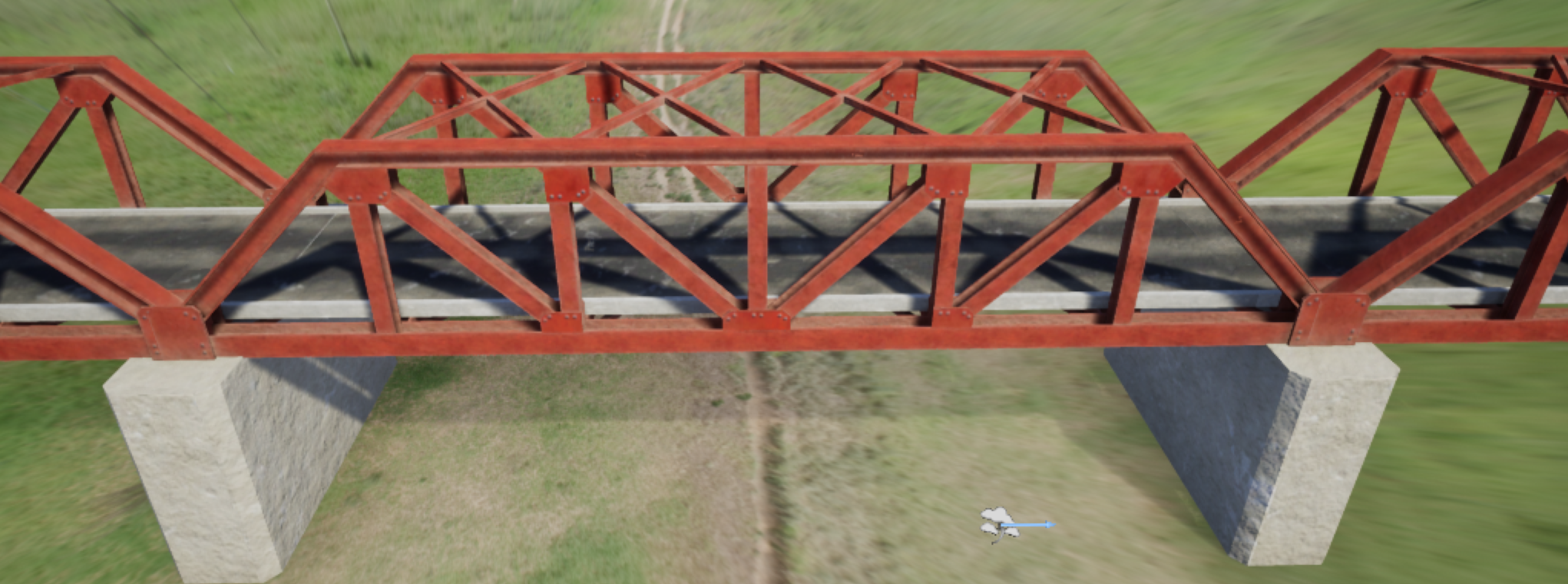}
        \caption{Iron Bridge}
    \end{subfigure}
    \begin{subfigure}[b]{0.49\columnwidth}
        \includegraphics[width = \textwidth]{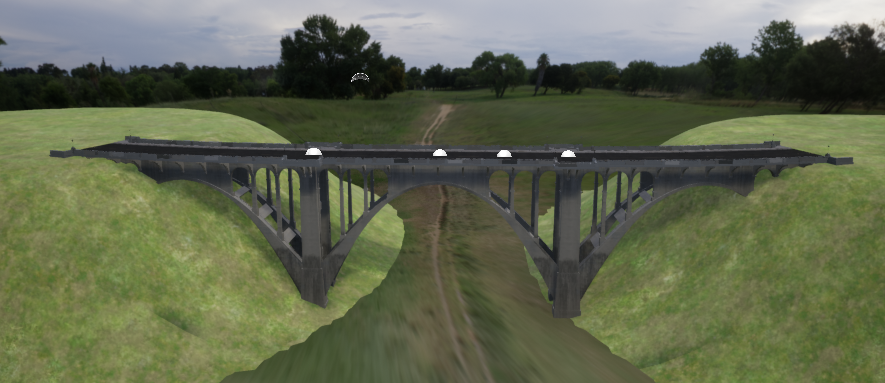}
        \caption{Steel Bridge}
    \end{subfigure}
    \caption{The 5 simulation environment bridges.}
    \label{fig:oriBridges}
\end{figure}

\subsubsection{Setup}
We use Robot Operating System (ROS) Melodic on Ubuntu 18.04 and AirSim to carry out the simulations. We equipped the simulated UAV with a LiDAR with specifications set to match a Velodyne VLP-16 3D LiDAR and an RGB camera. The 3D LiDAR generates around 300,000 points/sec. It also has a 360$^{\circ}$ horizontal field of view with $\pm$15$^{\circ}$ vertical field of view. The VLP-16 has a range of 100m~\cite{Velodyne67:online}. Due to the LiDAR's orientation, the LiDAR had a 30$^{\circ}$ horizontal FoV and measured distances ahead, behind, and below the UAV (with respect to the UAV's coordinate frame). We use the MoveIt~\cite{coleman2014reducing} software package based on the work done by Köse~\cite{tahsinko86:online} to implement the RRT connect algorithm. MoveIt uses RRT connect and the environmental 3D occupancy grid to find collision-free paths for point-to-point navigation.

For all experiments, we use a viewing cone with an apex angle of 0$^{\circ}$ and a viewing distance between two to ten meters. We set the viewing cone to the strictest possibility --- a straight line from the camera. This is to ensure the highest quality of images captured for inspection. We set the viewing distance based on Dorafshan et al.~\cite{dorafshan2017fatigue}, where they suggest a minimum flight distance of two meters to allow for the safe flight of the UAV.

\subsubsection{Qualitative Example}
\begin{figure}
    \centering
    \includegraphics[width = 0.75\columnwidth]{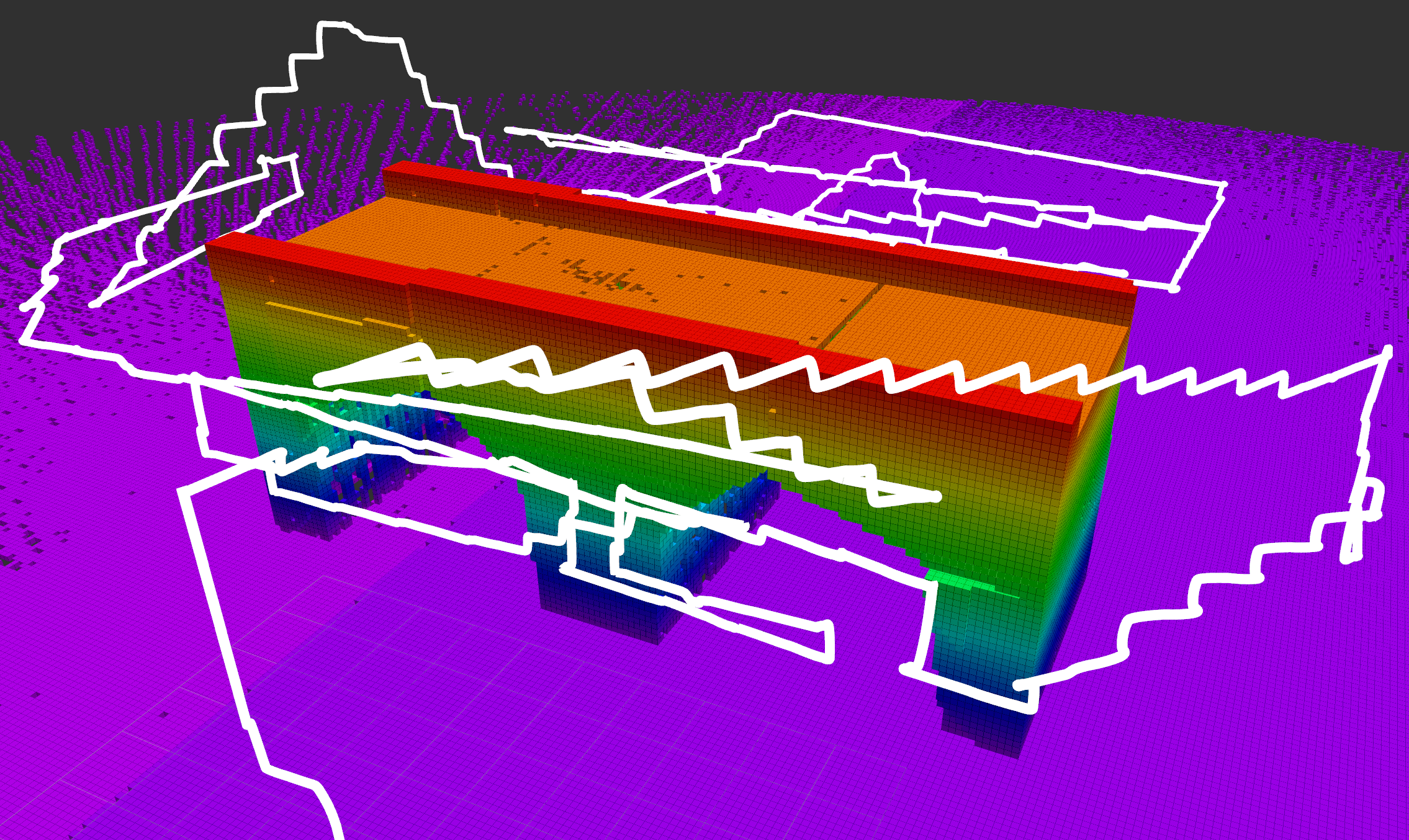}%
    \caption{UAV flight path during GATSBI.}%
    \label{fig:flightPath}%
\end{figure}

We evaluate GATSBI on five bridges shown in Fig.~\ref{fig:oriBridges}. The five bridges used are the arch bridge, box girder bridge, covered bridge, iron bridge, and steel bridge. All bridges, except the steel bridge, do not contain any other object in the environment except for the ground. The steel bridge, on the other hand, has other obstacles such as the landscape. They were all chosen because they distinctly represent different types of bridges. Figure~\ref{fig:flightPath} shows the path followed by the UAV as given by GATSBI around the Arch Bridge environment. We used AirSim's built-in segmentation algorithm to segment the bridge out from the rest of the environment. AirSim labels each LiDAR point with the segmentation color it belongs to. We modified the AirSim ROS wrapper to publish a segmented pointcloud containing points only belonging to the bridge using their built-in segmentation. The choice of the segmentation algorithm does not affect the main contribution of this paper which is the GATSBI planner. We also created a MoveIt implementation that integrates with AirSim. This package is available in our repository. Next, we present quantitative results on GATSBI.

\subsubsection{Computational Time}
\begin{figure}
    \centering
    \includegraphics[width = 0.75\columnwidth]{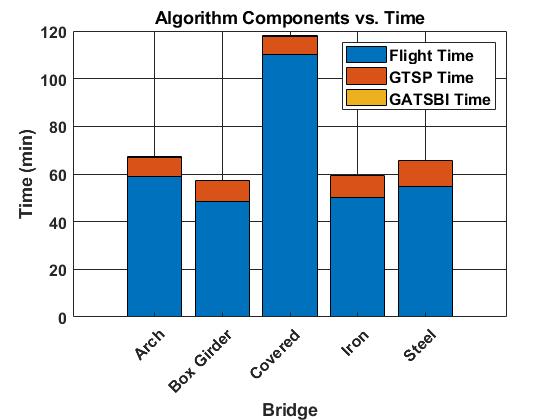}%
    \caption{We analyze how long different parts of the algorithm took to run. For all flights, the average flight time was 64.53 minutes. The average GTSP solver (GLNS) time was 8.82 minutes. Lastly, the average time for the remaining parts of GATSBI was 0.14 minutes.}%
    \label{fig:compTime}%
\end{figure}

We examine the time it takes for executing GATSBI. In Fig.~\ref{fig:compTime}, we report the average time for different components of the algorithm during all the simulations. We report three times: the time spent in the planner to create the GTSP instance (GATSBI), the time taken to solve the GTSP instance (GTSP), and the flight time before the algorithm calls the planner again (flight). We see that the time it takes GATSBI to perform segmentation and create a GTSP instance takes an average of 0.14 minutes. The GTSP solver takes an average of 8.82 minutes. Compared to the flight time (average of 64.5 minutes), the time taken by the planner is not significant. This suggests that GATSBI is not a bottleneck and is capable of running in real-time on UAVs that are executing 3D bridge inspection in unknown environments.

\subsubsection{Comparison with Baseline}\label{sec:sim:baseline}
\begin{figure}
    \centering
    \begin{subfigure}[b]{0.75\columnwidth}%
        \includegraphics[width = \textwidth]{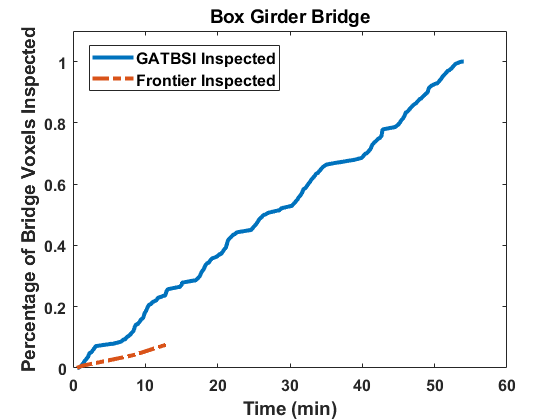}%
        % \label{fig:oriBridge}
    \end{subfigure}%
    \hfill%
    \begin{subfigure}[b]{0.75\columnwidth}%
        \includegraphics[width = \textwidth]{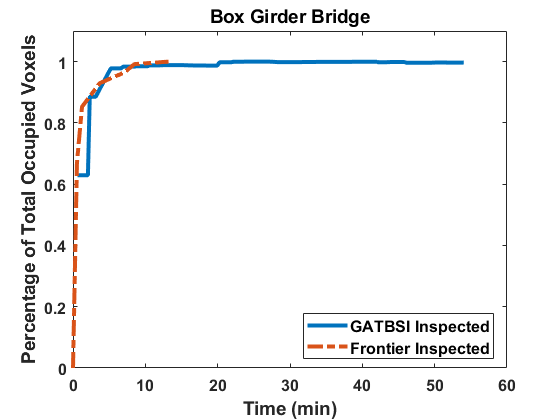}%
        % \label{fig:maskBridge}
    \end{subfigure}%
    \caption{Top: The solid blue line is the percentage of inspectable voxels inspected by GATSBI over time. The dashed orange line represents the same for Frontier Exploration. Bottom: The solid blue line represents the percentage of environment voxels discovered over time. The dashed orange represents the same for Frontier Exploration. While Frontier Exploration is faster at finding the environment voxels as seen on the bottom, it is not good at inspecting the bridge voxels and only inspects a small percentage of them.}
    \label{fig:results}
\end{figure}

We compare the performance of GATSBI with a baseline algorithm that we developed that is based on frontier exploration. 
\revone{Since the baseline method does not directly count the number of inspected voxels, we implement a package on top of the baseline to count the inspected voxels. This way we compare only the inspected voxels, not the covered voxels.} 
We can see in Fig.~\ref{fig:results} and Table~\ref{tab:results} that our method does better than the baseline method when comparing the percentage of bridge voxels inspected. We obtain this value by dividing the inspected bridge voxels ($|V_{BI}|$) upon the termination of the algorithm by the total inspectable bridge voxels. Note, obstructed bridge voxels that have no candidate viewpoints are uninspectable.

\begin{table}[ht!]
    \centering
    \begin{tabular}{|c|c|c|c|c|}
        \hline
        \textbf{Bridge} & \textbf{Algorithm} & \textbf{Bridge Vox.} & \textbf{Inspected} & \textbf{Runtime} \\
        \hline
        \multirow{2}{*}{Arch} & Frontier & \multirow{2}{*}{124} & 10 & 15.2 min \\
         & GATSBI & & 124 & 44.8 min \\
         \hline
        \multirow{2}{*}{Box Girder} & Frontier & \multirow{2}{*}{140} & 11 & 13.2 min \\
         & GATSBI & & 140 & 53.7 min \\
        \hline
        \multirow{2}{*}{Covered} & Frontier & \multirow{2}{*}{60} & 5 & 8.7 min \\
         & GATSBI & & 60 & 69.3 min \\
        \hline
        \multirow{2}{*}{Iron} & Frontier & \multirow{2}{*}{167} & 6 & 8.4 min \\
         & GATSBI & & 167 & 66.9 min \\
        \hline
        \multirow{2}{*}{Steel} & Frontier & \multirow{2}{*}{214} & 12 & 24.5 min \\
        & GATSBI & & 214 & 77.3 min \\
        \hline
    \end{tabular}
    \caption{Table showing the number of inspectable voxels that were inspected by GATSBI and Frontier for all 5 bridges as well as the total algorithm runtime.}
    \label{tab:results}
\end{table}

Nevertheless, we observe that GATSBI achieves inspection of 100\% voxels while the baseline only achieves a maximum of 10\%. Frontier exploration does not explicitly take inspection into account. As shown in the right plot in Fig.~\ref{fig:results}, it performs as well as GATSBI at exploration. The environment is also simple; in a more complicated environment, it would be better at exploration than the GATSBI algorithm. This validates our claim that GATSBI targets the inspection of bridge surfaces instead of just covering the environment. We also see that GATSBI executes a more thorough inspection than the baseline. Therefore, we justify the claim that GATSBI is more efficient in inspection compared to a frontier exploration algorithm.

\subsubsection{Parameter Tuning}
\begin{figure}
    \centering
    \begin{subfigure}[b]{0.75\columnwidth}
        \includegraphics[width = \textwidth]{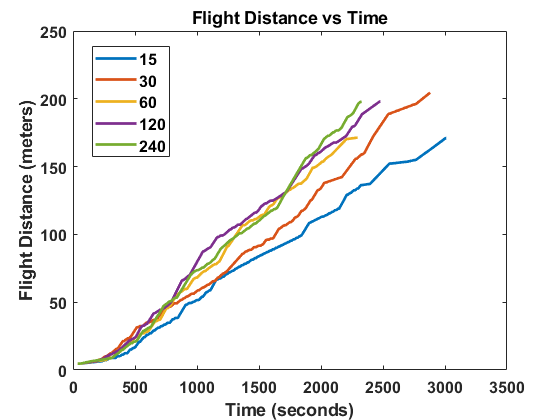}
        \caption{Replanning Time vs. Flight Distance and Total Runtime}
    \end{subfigure}
    \begin{subfigure}[b]{0.75\columnwidth}
        \includegraphics[width = \textwidth]{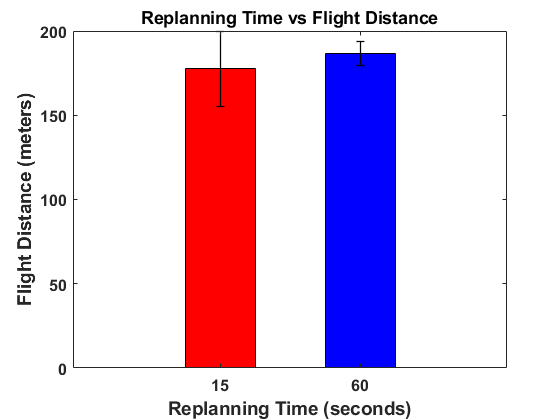}
        \caption{15 and 60 second Replanning Time vs. Flight Distance}
    \end{subfigure}
    \begin{subfigure}[b]{0.75\columnwidth}
        \includegraphics[width = \textwidth]{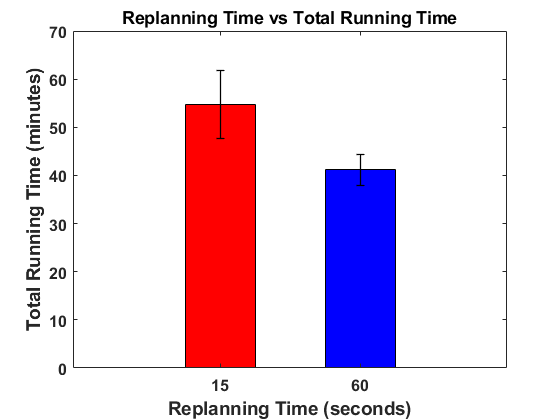}
        \caption{15 and 60 second Replanning Time vs. Total Runtime}
    \end{subfigure}
    \caption{Evaluation on different values for the replanning time parameter. In the first figure, we show how different values of replanning time affected the total flight distance of the UAV and the total runtime of the algorithm. For the second figure, we show the average and standard deviation of multiple runs at a replanning time of 15 and 60 seconds vs. the total flight distance. The last figure is the same but instead of total flight distance it is the total algorithm runtime.}
    \label{fig:replanning}
\end{figure}

One parameter used in the algorithm is replanning time, $RPT$. This time determines when to stop on the current GTSP tour if it has not been completed and replan with GATSBI using the most up-to-date environment and bridge information. Here, we discuss how we determined what to set $RPT$ to. Initially, we evaluated different values of $RPT$ for one of the bridge simulations. The results of this are shown in Fig.~\ref{fig:replanning}a. $RPT$ values of 15 and 60 seconds were then chosen for further evaluation due to them having the shortest flight distance. 5 runs each at these $RPT$ values were conducted using the simulation setup. Figure~\ref{fig:replanning}b shows the average flight distance during these 5 runs. On average, an $RPT$ value of 15 seconds had a slightly shorter total flight time (177.5 vs 186.5 meters) compared to an $RPT$ value of 60 seconds but with a higher standard deviation (22.4 vs 7.1). However, the average total runtime was longer using an $RPT$ value of 15 seconds (54.7 vs 41.2 min) while also having a higher standard deviation (7.0 vs 3.3) compared to an $RPT$ value of 60 seconds as shown in Fig.~\ref{fig:replanning}c. Because of this, an $RPT$ value of 60 seconds was used for the algorithm. 

\subsubsection{Crack Detection}

We also evaluated GATSBI on a simplified crack detection setup and compared its performance to frontier exploration. We randomly placed defects on 10\% of the bridge's surface voxels and determined how many of them were detected by both algorithms. Detected is defined as inspecting the voxel that the crack is on. This evaluation was done in an offline manner and each bridge was evaluated 10,000 times for both algorithms. The results are shown in Table~\ref{tab:cracks}. GATSBI performed at least 11.5x better at inspecting voxels with defects than frontier exploration. The reason that GATSBI's detection is not close to 100\% is that cracks were also placed on top of the bridge and below. For our bridge models, on average 40\% of the bridge surface is inspectable using a front-facing camera. Since the UAV is not equipped with an upwards and downwards-facing camera, the remaining 60\% cracks are not possible to be found. However, by adding these cameras and setting up the GTSP instance accordingly, these cracks would also be detected as long as they are not obstructed. Cracks could also be blocked by obstacles in the environment making them impossible to inspect.

\begin{table}[ht!]
    \centering
    \begin{tabular}{|c|c|c|c|}
        \hline
        \textbf{Bridge} & \textbf{Algorithm} & \textbf{\% Found} & \textbf{Std. Dev.} \\
        \hline
        \multirow{2}{*}{Arch} & Frontier & 2.6\% & 2.5\% \\
        & GATSBI & 32.5\% & 7.2\% \\
        \hline
        \multirow{2}{*}{Box Girder} & Frontier & 3.2\% & 2.8\% \\
        & GATSBI & 49.4\% & 8.0\% \\
        \hline
        \multirow{2}{*}{Covered} & Frontier & 3.0\% & 3.9\% \\
        & GATSBI & 34.5\% & 10.9\% \\
        \hline
        \multirow{2}{*}{Iron} & Frontier & 1.6\% & 1.9\% \\
        & GATSBI & 44.1\% & 7.6\% \\
        \hline
        \multirow{2}{*}{Steel} & Frontier & 2.4\% & 2.1\%\\
        & GATSBI & 42.8\% & 6.7\% \\
        \hline
    \end{tabular}
    \caption{Table showing the results from the crack detection simulations on the multiple bridges. We compare the percentage of cracks found using GATSBI and Frontier Exploration.}
    \label{tab:cracks}
\end{table}
\subsection{Real-World Experiments}

This section presents real-world experiments that evaluate the performance of GATSBI. Here, we show results with GATSBI since our simulated experiments demonstrated that GATSBI outperformed frontier exploration (our baseline in Section \ref{sec:sim:baseline}). Below, we present quantitative results that compare the computation time, flight time, and the number of voxels inspected with GATBSI. 

\subsubsection{Setup}
\begin{figure}[ht!]
    \centering
    \begin{subfigure}[b]{0.49\columnwidth}%
        \includegraphics[height = 3.25cm]{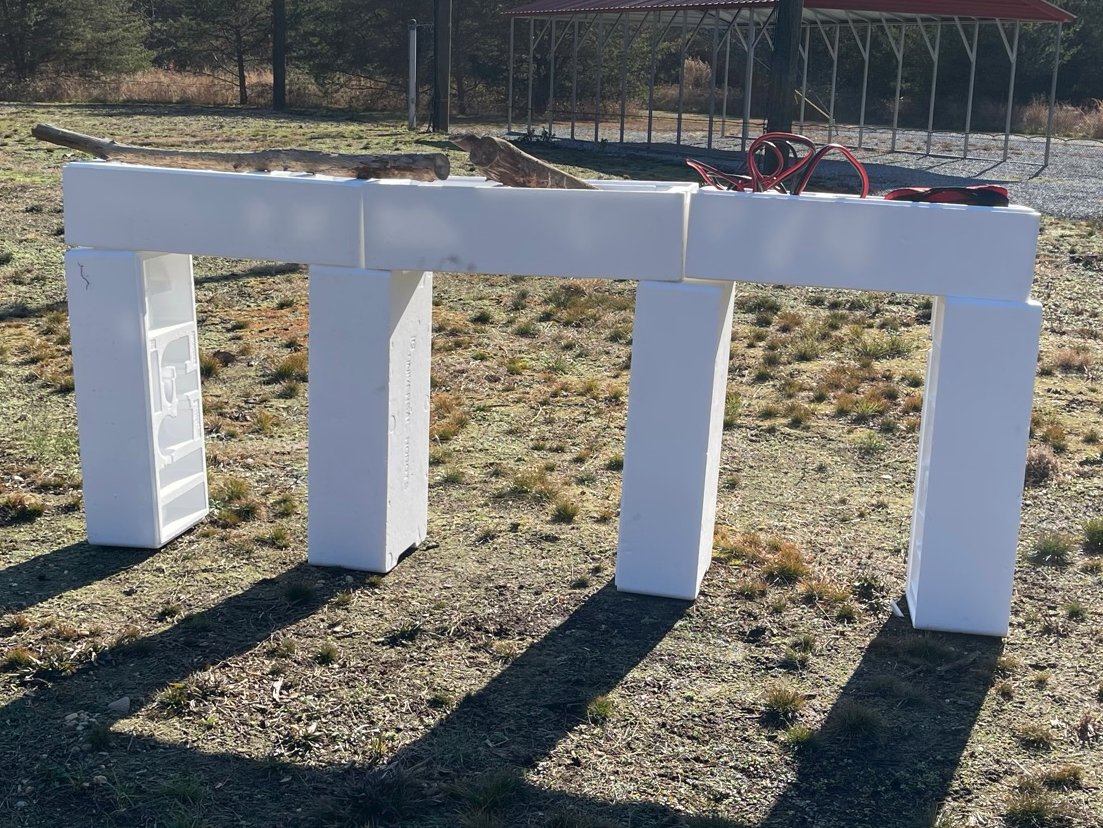}%
        %\label{fig:f3_setup}
    \end{subfigure}%
    \hfill%
    \begin{subfigure}[b]{0.49\columnwidth}%
        \includegraphics[height = 3.25cm]{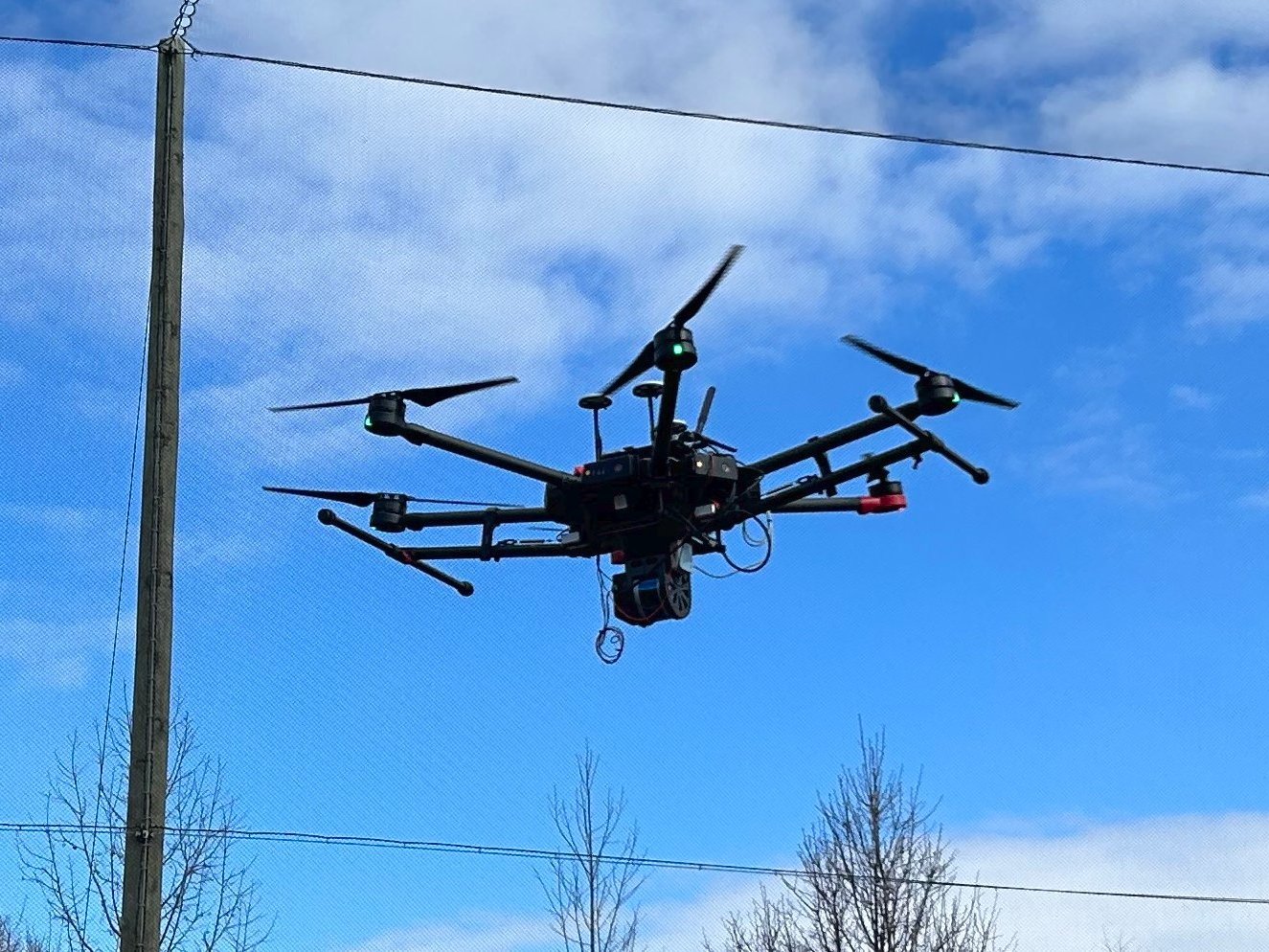}%
        %\label{fig:real_uav}
    \end{subfigure}%
    \caption{Left: The mock bridge used for experiments. Right: DJI M600 Pro mid-flight.}
    \label{fig:flight}
\end{figure}

We performed these experiments using a DJI Matrice M600 Pro (see Fig.~\ref{fig:flight}). The M600 Pro was equipped with an NVIDIA Jetson TX2 (which ran Ubuntu 16.04 and ROS Kinetic), Velodyne VLP-16 3D LiDAR, and GPS. Here, we used the same LiDAR model and parameters in the simulated experiments.

We constructed a mock bridge (Fig.~\ref{fig:flight}) and flew the UAV around it within an outdoor UAV cage called the Fearless Flight Facility (F3) at the University of Maryland, College Park. Due to mapping limitations, a simplified version of GATSBI was run. These limitations are addressed in Section~\ref{sec:con}. For our experiments, we used the UAV's LiDAR to create a map of our bridge and then ran GATSBI in an offline manner to find the inspection path. To localize the pointcloud, we used the pose of the DJI M600 Pro obtained using its GPS array. Unlike simulations where the localization is perfect, there is noise in the real world. To combat this, we can use off-the-shelf Visual Inertial Odometry along with GPS for localization along the bridge surface~\cite{kakillioglu20193d, song2019method}. For our experiments, the bridge was segmented out using a box filter on the localized pointcloud. For our use-case, geographical segmentation worked well but for more complicated experiments, color-based or learning-based segmentation networks can be used. Another implementation of MoveIt was integrated with DJI's SDK and used for our obstacle-avoidance navigation planning. This package is also provided in our repository and is the first of its kind. For larger structures, battery-life of the UAV would limit structure coverage during the single flight. We can account for this by returning the UAV to the home-point when it's battery gets low and swapping out the batteries with charged ones. In the future, we are interested in investigating how GATSBI scales when we use multiple agents to inspect for bridge.

\begin{figure}[ht!]
    \centering
    \includegraphics[width = 0.7\columnwidth]{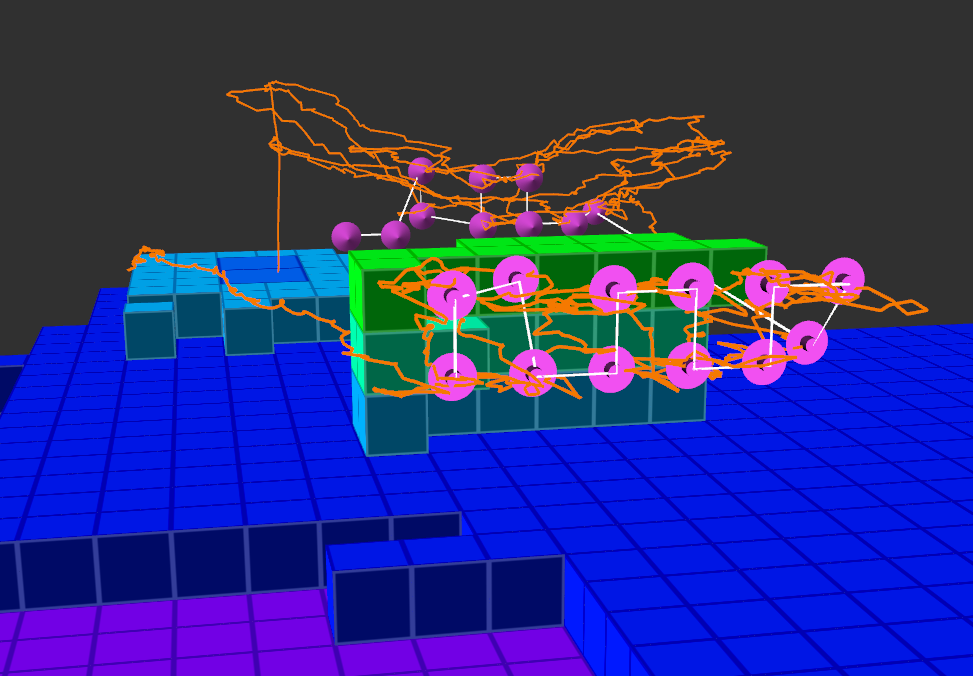}%
    \caption{Flight path and target inspection path on the real-world mock bridge.}%
    \label{fig:paths_hardware}%
\end{figure}

\subsubsection{Results}
We implemented a simplified version of the GATSBI algorithm on the real-world mock bridge. Our algorithm was able to inspect all 15 inspectable bridge voxels. The target inspection path and actual flown path are shown in Fig.~\ref{fig:paths_hardware}. The white line is the direct path between the target inspection points. The orange line is the actual flown path. The pink cones represent the inspection points. This experiment validates that GATSBI can be used for real-world infrastructure inspection. The computation time of the GATSBI algorithm was 0.32 seconds, GTSP time was 8.25 seconds, and flight time was 772 seconds. As shown in the simulations, the GATSBI algorithm time is not the bottleneck. Including the GTSP solver time, the total algorithm time is still much smaller than the flight time.

%
%\begin{table}[ht!]
%    \centering
%    \begin{tabular}{|c|c|}
%        \hline
%        \textbf{Bridge Vox.} & \textbf{Inspected} \\
%        \hline
%        15 & 15 \\
%        \hline
%    \end{tabular}
%    \caption{Table showing the number of inspectable voxels that were inspected by GATSBI during real-world experiment.}
%    \label{tab:results_experiment}
%\end{table}

%\begin{table}[ht!]
%    \centering
%    \begin{tabular}{|c|c|c|}
%        \hline
%        \textbf{GATSBI Time} & \textbf{GTSP Time} & \textbf{Flight Time}\\
%        \hline
%       0.32s & 8.25s & 772s \\
%        \hline
%    \end{tabular}
%    \caption{Table showing the time taken for each part of GATSBI during hardware experiment}
%    \label{tab:results_experiment_comp}
%\end{table}

\section{Conclusion}\label{sec:con}
We present GATSBI, a 3D bridge inspection planner. We evaluate the performance of the algorithm through AirSim simulations and real-world hardware experiments with a UAV equipped with a 3D LiDAR and an RGB camera. The simulations show that GATSBI outperforms a frontier-based exploration algorithm. The hardware experiments show that GATSBI is a viable solution to real-world infrastructure inspection. In particular, we show that the algorithm is efficient in the sense that it targets inspectable voxels rather than simply exploring a volume. The simulations and experiments also demonstrate that the algorithm can run in real-time. In future work, we intend to improve our real-world experiments. In particular, implementing a SLAM module to account for noise found in the real world is needed to conduct more complex experiments. We are also looking into implementing a multi-agent solution to account for limited battery-life of UAV's.

\bibliographystyle{unsrt}
\bibliography{main.bib}

\end{document}